%% file: vizzo2023ral.tex
\documentclass[letterpaper, 10 pt, journal, twoside]{IEEEtran}

\usepackage{hyperref}

\input{stachnisslab-latex}

\input{stachnisslab-math}

\input{stachnisslab-custom}

\input{stachnisslab-glossary}

\makeatletter
\def\blfootnote{\gdef\@thefnmark{}\@footnotetext}
\makeatother
\begin{document}

\title{KISS-ICP: In Defense of Point-to-Point ICP -- \\ Simple, Accurate, and Robust Registration \\If Done the Right Way}
\renewcommand{\and}{\hspace{0.5cm}}
\author{Ignacio Vizzo \and Tiziano Guadagnino \and Benedikt Mersch \and Louis Wiesmann \and Jens Behley \and Cyrill Stachniss%
}

\twocolumn[{%
			\renewcommand\twocolumn[1][]{#1}%
			\maketitle
			\begin{center}
				\vspace{-12mm}
				\includegraphics[width=0.91\textwidth]{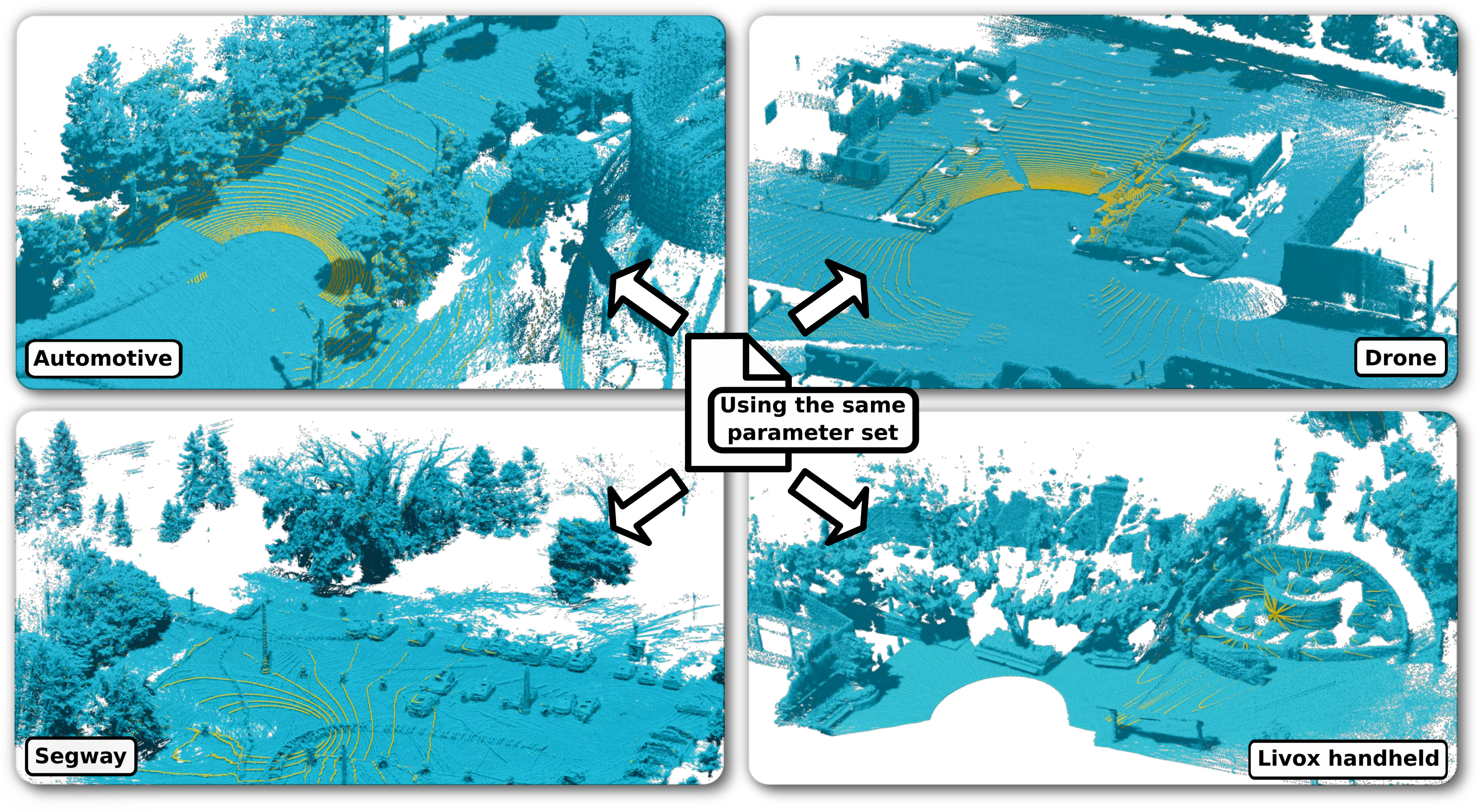}
				\captionof{figure}{Point cloud maps (blue) generated by our proposed odometry pipeline on different datasets with the same set of parameters. We depict the latest scan in yellow. The scans are recorded using different sensors with different point densities, different orientations, and different shooting patterns. The automotive example stems from the \mulran. The drone of the Voxgraph dataset~\cite{reijgwart2019ral} and the segway robot used in the \nclt show a high acceleration motion profile. The handheld Livox LiDAR~\cite{lin2019iros-larl} has a completely different shooting pattern than the commonly used rotating mechanical LiDAR.}
				\label{fig:motivation}
			\end{center}%
		}]

\markboth{IEEE Robotics and Automation Letters. Preprint Version. Accepted December, 2022.}
{Vizzo \MakeLowercase{\textit{et al.}}: KISS-ICP: In Defense of Point-to-Point ICP -- Simple, Accurate, and Robust Registration If Done the Right Way}

\begin{abstract}
	Robust and accurate pose estimation of a robotic platform, so-called sensor-based odometry, is an essential part of many robotic applications. While many sensor odometry systems made progress by adding more complexity to the ego-motion estimation process, we move in the opposite direction. By removing a majority of parts and focusing on the core elements, we obtain a surprisingly effective system that is simple to realize and can operate under various environmental conditions using different LiDAR sensors.
	Our odometry estimation approach relies on point-to-point ICP combined with adaptive thresholding for correspondence matching, a robust kernel, a simple but widely applicable motion compensation approach, and a point cloud subsampling strategy. This yields a system with only a few parameters that in most cases do not even have to be tuned to a specific LiDAR sensor.
	Our system performs on par with state-of-the-art methods under various operating conditions using different platforms using the same parameters: automotive platforms, UAV-based operation, vehicles like segways, or handheld LiDARs. We do not require integrating IMU data and solely rely on 3D point clouds obtained from a wide range of 3D LiDAR sensors, thus, enabling a broad spectrum of different applications and operating conditions.
	Our open-source system operates faster than the sensor frame rate in all presented datasets and is designed for real-world scenarios.
\end{abstract}


\blfootnote{Manuscript received: September 14, 2022; Revised: December 5, 2022; Accepted: December 27, 2022. This paper was recommended for publication by Editor Javier Civera upon evaluation of the Associate Editor and Reviewers' comments. \\ \indent This work has partially been funded by the Deutsche Forschungsgemeinschaft (DFG, German Research Foundation) under Germany's Excellence Strategy, EXC-2070 -- 390732324 -- PhenoRob and by the European Union’s HORIZON research and innovation programme under grant agreement No~101070405~(Digiforest). \\ \indent All authors are with the University of Bonn, Germany. Cyrill Stachniss is additionally with the Department of Engineering Science at the University of Oxford, UK, and with the Lamarr Institute for Machine Learning and Artificial Intelligence, Germany. \\ \indent Digital Object Identifier (DOI): see top of this page.}%

\section{Introduction}
\label{sec:intro}
\IEEEPARstart{O}{dometry} estimation is an essential building block for any mobile robot that needs to autonomously navigate in unknown environments. In the LiDAR sensing domain, current odometry pipelines typically use some form of~\ac*{icp} to estimate poses incrementally~\cite{dellenbach2022icra,serafin2015iros,vizzo2021icra,zhang2014rss}. Even though LiDAR odometry has been an active area of research for the last three decades, the design of current systems is usually coupled with assumptions about the robot motion~\cite{dellenbach2022icra} and the structure of the environment~\cite{shan2018iros} to achieve accurate and robust alignment results. To the best of our knowledge, no existing 3D LiDAR odometry approach is free of parameter tuning and works out of the box in different scenarios, using arbitrary LiDAR sensors, supporting different motion profiles, and consequently types of robots, such as ground and aerial robots.

This paper returns to the roots: classical point-to-point~\ac*{icp}, introduced~30 years ago by Besl and McKay~\cite{besl1992pami}. We aim to tackle the inherent problems of sequentially operating LiDAR odometry systems that prohibit current approaches from generalizing to different environments, sensor resolutions, and motion profiles using a single configuration.
We present simple yet effective reasoning about the robot kinematics and the sequential way LiDAR data is recorded on a mobile platform, as well as an effective downsampled point cloud representation that allows us to minimize the need for parameter tuning.

Our system challenges even extensively hand-tuned and optimized existing~\ac*{slam} systems. Our design uses neither sophisticated feature extraction techniques, learning methods, nor loop closures. The same parameter set works in various challenging scenarios such as highway drives of robot cars with many dynamic objects, drone flights, handheld devices, segways, and more. Thus, we take a step back from mainstream research in LiDAR odometry estimation and focus on reducing the components to their essentials. This makes our system perform extraordinarily well in various real-world scenarios, see~\figref{fig:motivation}.

The main contribution of this paper is a simple yet highly effective approach for building LiDAR odometry systems that can accurately compute a robot's pose online while navigating through an environment. We identify the core components and properly evaluate the impact of different modules on such systems.
We show that with the proper use of~\ac*{icp} that builds on basic reasoning about the system's physics and the sensor data's nature, we obtain competitive odometry. Besides motion prediction, spatial scan downsampling, and a robust kernel, we introduce an adaptive threshold approach for~\ac*{icp} in the context of robot motion estimation that makes our approach effective and, at the same time, generalizes easily.

We make three key claims: Our ``keep it small and simple'' approach exploiting point-to-point~\ac*{icp} is
(i) on par with state-of-the-art odometry systems,
(ii) can accurately compute the robot's odometry in a large variety of environments and motion profiles with the same system configuration, and
(iii) provides an effective solution to motion distortion without relying on IMUs or wheel odometers.
In sum, ``good old point-to-point~\ac*{icp}'' is a surprisingly powerful tool, and there is little need to move to more sophisticated approaches if the basic components are done well.

We provide an open-source implementation at:~\url{https://github.com/PRBonn/kiss-icp} that precisely follows the description of this paper.

\section{Related Work}
\label{sec:related}
Point cloud registration has been an active area of research for the last three decades~\cite{besl1992pami,chen1991iros} and is still relevant nowadays. The~\ac*{icp} algorithm can solve the problem of finding a transformation that brings two different point clouds into a common reference frame, and it is a special case of the absolute orientation problem in photogrammetry.~\ac*{icp} typically consists of two parts. The first one is to find correspondences between the point clouds. The second one computes the transformation that minimizes an objective function defined on the correspondences from the first step. One repeats this process until a convergence criterion is met. Most~\ac*{icp} variants~\cite{behley2018rss, dellenbach2022icra, deschaud2018icra, serafin2015iros, pan2021icra-mvls,zhang2014rss} utilize a maximum distance threshold in the data association module plus a robust kernel~\cite{chebrolu2021ral} and a maximum number of iterations. In contrast, we propose a threshold estimation method that adapts to changing scenarios by reasoning about the system kinematics and the nature of the data in combination with a robust kernel. We avoid controlling the number of iterations of the~\ac*{icp} to achieve better generalization.

\ac*{icp} can be used to obtain an odometry estimation from streaming data from a sensor such as \rgbd cameras~\cite{newcombe2011ismar} or LiDARs~\cite{dellenbach2022icra}. In this work, we focus on the problem of LiDAR odometry estimation, although the ideas presented can be easily extended to other range-sensing technologies.

Nearly all modern~\ac*{slam} systems build on top of odometry algorithms. Zhang~\etalcite{zhang2014rss} proposed~\ac*{loam} that computes the robot's odometry by registering planar and edge features to a sparse feature map.~\ac*{loam} inspired numerous other works~\cite{shan2020iros,wang2021iros-fflo}, such as Lego-LOAM~\cite{shan2018iros}, which adds ground constraints to improve accuracy, and recently F-LOAM~\cite{wang2021iros-fflo}, which revised the original method with a more efficient optimization technique enabling faster operation. However, these methods rely on hand-tuned feature extraction, which typically requires tedious parameter tuning that depends on sensor resolution, environment structure,  etc.~In contrast, we only rely on point coordinates removing this data-dependent parameter adaptation.

Behley and Stachniss~\cite{behley2018rss} propose the surfel-based method SuMa to achieve LiDAR odometry estimation and mapping. It has also been extended to account for semantics~\cite{chen2019iros} and explicitly handle dynamic objects~\cite{chen2021ral}. In contrast to the surfel-based mapping, Deschaud~\cite{deschaud2018icra} introduced \imls selecting an implicit moving least square surface~\cite{kolluri2008talg} as map representation. Along these lines, Vizzo~\etalcite{vizzo2021icra} exploited a triangular mesh as the internal map representation. All the above approaches rely on a point-to-plane~\cite{rusinkiewicz2001dim} metric to register consecutive scans. This requires normal estimation, which introduces additional data-dependent parameters. Furthermore, noisy 3D information can impact the normal computation and subsequently the registration in a negative way. We will show that by minimizing a simpler point-to-point metric, we obtain on-par or better odometry performance. Moreover, this design choice enables us to represent the internal map as a voxelized, downsampled point cloud, simplifying the implementation.

Recently, several new approaches~\cite{dellenbach2022icra,pan2021icra-mvls,shan2020iros} have been proposed to solve the odometry estimation problem. Most of these works focus on the runtime operation of the system as well as on the accuracy. Pan~\etalcite{pan2021icra-mvls} propose a multi-metric system~(MULLS) that obtains good results in many challenging scenarios at the cost of tuning many parameters for each run. Dellenbach~\etalcite{dellenbach2022icra} introduced a novel approach, called~\ac*{cticp}, which incorporates the motion un-distortion into the registration showing great results but adding more complexity. Additionally, the robots' motion profile must be known a priori, as, for example, a car will have a different profile than a segway platform. We challenge the need for sophisticated optimization techniques to cope with motion distortion requiring only the constant velocity model. Furthermore, our system only relies on a few parameters, and we do not need to know the motion profile in advance.

Many state-of-the-art odometry systems~\cite{behley2018rss,dellenbach2022icra,pan2021icra-mvls,shan2020iros} also rely on pose graph optimization to achieve a better alignment. In contrast, we do not exploit such techniques and state that pose graph optimization is orthogonal to the presented approach and can be easily integrated.
In sum, we step back from the common mainstream work on LiDAR odometry and propose a system that solely relies on a point-to-point metric and does not employ pose graph optimization~\cite{behley2018rss,dellenbach2022icra,pan2021icra-mvls,shan2020iros}. Our system can run on different types of mobile robots, drones, handheld devices, and segways, without the need to fine-tune the system to a specific application.

\section{KISS-ICP -- Keep It Small and Simple}
\label{sec:main}

This work aims to incrementally compute the trajectory of a moving LiDAR sensor by sequentially registering the point clouds recorded by the scanner. We reduced the components to a minimal set needed to build an effective, accurate, robust, and still reasonably simple LiDAR odometry system.

For each 3D scan in form of a local, egocentric point cloud~$\set{P}\,{=}\,\{ \d{p}_i \,{\mid}\, \d{p}_i \,{\in}\, \RR^3 \}$, we perform the following four steps to obtain a global pose estimate~\mbox{$\mq{T}_t\,{\in}\,SE(3)$} at time~$t$. First, we apply sensor motion prediction and motion compensation, often called deskewing, to undo the distortions of the 3D data caused by the sensor's motion during scanning. Second, we subsample the current scan. Third, we estimate correspondences between the input point cloud and a reference point cloud, which we call the local map. We use an adaptive thresholding scheme for correspondence estimation, restricting possible data associations and filtering out potential outliers. Fourth, we register the input point cloud to the local map using a robust point-to-point ICP algorithm. Finally, we update the local map with a downsampled version of the registered scan. Below, we describe these components in detail.

\subsection{Step~1: Motion Prediction and Scan Deskewing}
\label{sec:velocity}
We advocate for rethinking the point cloud registration in the context of mobile robots, which continuously record data. One should not think of it as registering arbitrary pairs of 3D point clouds. Instead, one should phrase it as \textit{estimating how much the robot's actual motion deviates from its expected motion by registering consecutive scans}.

Different approaches can be used to compute the robot's expected motion before considering the LiDAR data. The three most popular choices are the constant velocity model, wheel odometry obtained through encoders, and IMU-based motion estimation. The constant velocity~\cite{thrun2005probrobbook} model assumes that a robot moves with the same translational and rotational velocity as in the previous time step. It requires no additional sensors (no wheel encoder, no IMU) and thus is the most widely applicable option.

Our approach uses the constant velocity model for two reasons: first, it is generally applicable, requires no additional sensors, and avoids the need for time synchronization between sensors. Second, as we will show in our experimental evaluation, it works well enough to provide a solid initial guess when searching for data associations and deskewing 3D scans. This follows from the fact that robotic LiDAR sensors commonly record and stream point clouds at 10\,Hz to 20\,Hz, i.e., every 0.05\,s to 0.1\,s. In most cases, the acceleration or deceleration, \ie, the deviations from the constant velocity model that occurs within such short time intervals, are fairly small. If the robot accelerates or decelerates, the constant velocity estimation of the robot's pose will be slightly off, and therefore, we need to correct this estimate through registration. These accelerations determine the possible displacements of the (static) 3D points.

The constant velocity model approximates the translational and angular velocities, denoted as~$\d{v}_t$ and~$\d{\omega}_t$ at time~$t$ respectively, by using the previous pose estimates~$\mq{T}_{t-1} \,{=}\, (\m{R}_{t{-}1},\d{t}_{t{-}1})$ and~$\mq{T}_{t-2} \,{=}\, (\m{R}_{t-2}, \d{t}_{t-2})$, represented by a rotation matrix $\m{R}_t \,{\in}\, SO(3)$ and a translation vector~$\d{t}_t \,{\in}\, \RR^3$ for the time step~$t$. We first compute the relative pose $\mq{T}_{\text{pred}, t}$ that we will use as motion prediction as:
\begin{equation}
	\mq{T}_{\text{pred}, t} = \begin{bmatrix}\m{R}_{t-2}^\top \,\m{R}_{t-1} & \m{R}_{t-2}^\top\,(\d{t}_{t-1} - \d{t}_{t-2}) \\
               \m{0}                          & 1\end{bmatrix},
\end{equation}
then derive the corresponding velocities as:
\begin{align}
	\d{v}_{t}      & = \frac{\m{R}_{t-2}^\top\,(\d{t}_{t-1} - \d{t}_{t-2})}{\Delta t}, \\
	\d{\omega}_{t} & = \frac{\text{Log}(\m{R}_{t-2}^\top \,\m{R}_{t-1})}{\Delta t},
\end{align}
where $\Delta t$ is the acquisition time of one LiDAR sweep, typically 0.05\,s or 0.1\,s, and Log:~$SO(3) \, {\to} \, \RR^3$ extracts the axis-angle representation.

Note that also wheel odometry or an IMU-based motion prediction approach can be used instead to compute $\d{v}_t$ and $\d{\omega}_t$ for each time step. This will not change the remainder of our approach. For example, if one has good wheel odometry available, this can also be used. However, we use constant velocity as a generally applicable approach.

Within the acquisition time~$\Delta t$ of one LiDAR sweep, multiple 3D points are measured by the scanner. The relative timestamp~$s_i  \, {\in} \, [0,\Delta t]$ for each point \mbox{$\d{p}_i \, {\in} \, \set{P}$} describes the recording time relative to the scan's first measurement. This relative timestamp allows us to compute the motion compensation resulting in a deskewed point~\mbox{$\d{p}_i^{*} \, {\in} \, \set{P}^*$} of the corrected scan $\set{P}^*$ reading by
\begin{align}
	\label{eq:deskew}
	\d{p}_i^{*} & = \text{Exp}(s_i\d{\omega}_{t}) \d{p}_i + s_i\d{v}_{t},
\end{align}
where Exp:~$\RR^3 \, {\to} \, SO(3)$ computes a rotation matrix from an axis-angle representation. Note that~$\text{Exp}(s_i\d{\omega}_{t})$ is equivalent to performing SLERP in the axis-angle domain.

This form of scan deskewing, especially with the constant velocity model, is easy to implement, generally applicable, and does not require additional sensors, high-precision time synchronization between sensors, or IMU biases to be estimated. As we show in \secref{sec:exp}, this approach often performs even better than more complex compensation systems~\cite{dellenbach2022icra}, at least as long the motion between the start and end of the sweep is small as it is for most robotics applications.

\subsection{Step 2: Point Cloud Subsampling}
\label{sec:sampling}

Identifying a set of keypoints in the point cloud is a common approach for scan registration~\cite{guadagnino2022ral,rusinkiewicz2001dim, zhang2014rss}. It is typically done to achieve faster convergence and/or higher robustness in the data association. However, complex filtering of the point cloud usually comes with an extra layer of complexity and parameters that often need to be tuned.

Rather than extracting 3D keypoints, which often requires environment-dependent parameter tuning, we propose to compute only a spatially downsampled version~$\set{\hat{P}}^*$ of the deskewed scan~$\set{P}^*$. Downsampling is done using a voxel grid. As we will explain in~\secref{sec:target-map} below in more detail, we use a voxel grid as our local map, where each voxel call has a size of $v \,{\times}\, v \,{\times}\, v$ and each cell only store a certain number of points. Every time we process an incoming scan, we first downsample the point cloud of the scan to an intermediate point cloud $\set{P}^*_\text{merge}$, which is later used to update the map when the relative motion of the robot has been determined with ICP. To obtain the points in $\set{P}^*_\text{merge}$, we use voxel size $\alpha \, v$ with $\alpha \,{\in}\, (0.0,1.0]$ and keep only a single point per voxel.

For the ICP registration, an even lower resolution scan is beneficiary. Thus, we compute a further reduced point cloud~$\set{\hat{P}}^*$ by downsampling $\set{P}^*_\text{merge}$ again using a voxel size of $\beta \, v$ with $\beta \,{\in}\, [1.0,2.0]$ keeping only a single point per voxel. This further reduces the number of points processed during the registration and allows for a fast and highly effective alignment. The idea of this ``double downsampling'' stems from~\ac*{cticp}~\cite{dellenbach2022icra}, the so far best performing open-source LiDAR odometry system on KITTI.

Most voxelization approaches, however, select the center of each occupied voxel to downsample the point cloud~\cite{rusu2011icra,zhou2018arxiv}. Instead, we found it advantageous to maintain the original point coordinates, select only one point per voxel for a single scan, and keep its coordinates to avoid discretization errors. This means the reduced cloud is a subset of the deskewed one, \ie, \mbox{$\set{\hat{P}}^* \,{\subseteq}\, \set{P}^*$}. In our implementation, we keep only the first point that was inserted into the voxel.

\subsection{Step 3: Local Map and Correspondence Estimation}
\label{sec:target-map}

In line with prior work~\cite{behley2018rss,dellenbach2022icra,newcombe2011ismar,zhang2014rss}, we register the deskewed and subsampled scan $\set{\hat{P}}^*$ to the point cloud built so far, \ie, a local map, to compute an incremental pose estimate~$\Delta \mq{T}_{\text{icp}}$. We use frame-to-map registration as it proves more reliable and robust than the frame-to-frame alignment~\cite{behley2018rss,newcombe2011ismar}. To do that effectively, we must define a data structure representing the previously registered scans.

Modern approaches have used very different types of representations for this local map. Popular approaches are voxel grids~\cite{zhang2014rss}, triangle meshes~\cite{vizzo2021icra}, surfel representations~\cite{behley2018rss}, or implicit representations~\cite{deschaud2018icra}. As mentioned in \secref{sec:sampling}, we utilize a voxel grid to store a subset of 3D points. We use a grid with a voxel size of $v\,{\times}\,v\,{\times}\,v$ and store up to $N_{\text{max}}$ points per voxel. After registration, we update the voxel grid by adding the points~$\{\mq{T}_t\,\d{p} \,{\mid}\, \d{p} \,{\in}\, \set{P}^*_\text{merge}\}$ from the new scan using the global pose estimate $\mq{T}_t$. Voxels that already contain $N_{\text{max}}$ points are not updated. Additionally, given the current pose estimate, we remove voxels outside the maximum range $r_\text{max}$. Thus, the size of the map will stay bounded.

Instead of a 3D array, we use a hash table to store the voxels, allowing a memory-efficient representation and fast nearest neighbor search~\cite{dellenbach2022icra,niessner2013siggraph}. However, the used data structure can be easily replaced with VDBs~\cite{museth2013siggraph,vizzo2022sensors}, Octrees~\cite{vespa2018ral,zeng2013gmodels}, or KD-Trees~\cite{bentley1975kdtree}.

\subsection{Adaptive Threshold for Data Association}
\label{sec:adaptive}
\ac*{icp} typically performs a nearest neighbor data association to find corresponding points between two point clouds~\cite{besl1992pami}. When searching for associations, it is common to impose a maximum distance between corresponding points, often using a value of 1\,m or 2\,m~\cite{behley2018rss,vizzo2021icra,zhang2014rss}.
This maximum distance threshold can be seen as an outlier rejection scheme, as all correspondences with a distance larger than this threshold are considered outliers and are ignored.

The required value for this threshold $\tau$ depends on the expected initial pose error, the number and type of dynamic objects in the scene, and, to some degree, the sensor noise. It is typically selected heuristically. Based on the considerations about the constant velocity motion prediction in \secref{sec:velocity}, we can, however, estimate a likely limit from data by analyzing how much the odometry may deviate from the motion prediction over time. This deviation $\Delta \mq{T}$ in the pose corresponds exactly to the local~\ac*{icp} correction to be applied to the predicted pose (but it is not known beforehand). Intuitively, we can observe the robot's acceleration in the magnitude of $\Delta \mq{T}$. If the robot is not accelerating, then $\Delta \mq{T}$ will have a small magnitude, often around zero, meaning that the constant velocity assumption holds and no correction has to be done by~\ac*{icp}.

We integrate this information into our data association search by exploiting the so-far successful~\ac*{icp} executions. We can estimate the possible point displacement between corresponding points in successive scans in the presence of a potential acceleration expressed through $\Delta \mq{T}$ as:
\begin{equation}
	\label{eq:delta-points}
	\delta(\Delta \mq{T}) = \delta_{\text{rot}}(\Delta \m{R}) + \delta_{\text{trans}}(\Delta \d{t}),
\end{equation}
where $\Delta \m{R} \,{\in}\, SO(3)$ and $\Delta \d{t} \,{\in}\, \RR^{3}$ refer to the rotational and translational component of the deviation, given by
\begin{align}
	\label{eq:deltaT}
	\delta_{\text{rot}}(\Delta \m{R})   & =  2 \ r_\text{max} \sin\Biggl(\frac{1}{2} \underbrace{\arccos\left(\frac{\Tr(\Delta \m{R})-1}{2}\right)}_{\theta}\Biggl) \\
	\delta_{\text{trans}}(\Delta \d{t}) & = \|\Delta \d{t}\|_{2}                                                                             .
\end{align}

\begin{figure}
	\centering
	\def\svgwidth{0.7\linewidth}
	\fontsize{9}{9}\selectfont
	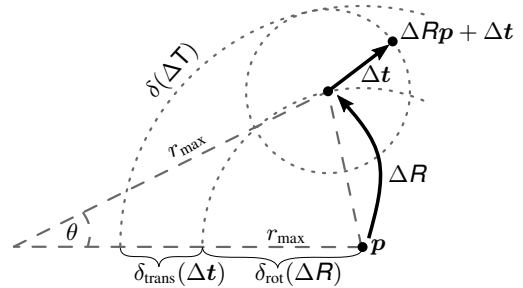
	\caption{Exemplary computation of the maximum point displacement $\delta(\Delta \mq{T})$ caused by a rotational and translational deviation $(\Delta \m{R}, \Delta\d t)$ from the predicted motion.}
	\label{fig:adaptive}
	\vspace{-0.6cm}
\end{figure}

The term $\delta_{\text{rot}}(\Delta \m{R})$ represents the displacement that occurs for a range reading with maximum range $r_\text{max}$ subject to the rotation $\Delta \m{R}$, see also~\figref{fig:adaptive}. Note that~\eqref{eq:delta-points} constitutes an upper bound for the point displacement as
\begin{equation}
	\|\Delta \m{R}\,\d{p}+\Delta \d{t}-\d{p}\|_2 \leq \delta_{\text{rot}}(\Delta \m{R}) + \delta_{\text{trans}}(\Delta \d{t}),
\end{equation}
which follows from the triangle inequality.

For obtaining $\delta_{\text{rot}}$ other approaches could be considered, like taking into account the individual ranges for the adaptive threshold computation~\cite{blanco2014mrpt}. In our tests, we did not see any difference in the results but a 3-fold increase of the overall runtime; thus, we use $r_{\text{max}}$ instead of $r$ for computing $\delta_{\text{rot}}$.

To compute the threshold $\tau_t$ at time~$t$, we consider a Gaussian distribution over~$\delta $ using the values of~\eqref{eq:delta-points} over the trajectory computed so far whenever the deviation was larger than a minimum distance $\delta_\text{min}$, i.e., situations where the robot's motion was deviating from the constant velocity model. Its standard deviation is
\begin{align}
	\label{eq:threshold}
	\sigma_t & = \sqrt{ \frac{1}{|\mathcal{M}_t|} \sum_{i \in \mathcal{M}_t}  \delta(\Delta T_{i})^2},
\end{align}
where the index set $\mathcal{M}_t$ of deviations up to~$t$ is given by
\begin{align}
	\mathcal{M}_t & = \{i \mid i < t \wedge \delta(\Delta T_{i}) > \delta_\text{min} \}.
\end{align}

This avoids reducing the value of $\sigma_{t}$ too much when the robot is not moving or is moving at constant velocity for a long time. In our experiments, we set this threshold $\delta_{\min}$ to~0.1\,m. We then compute the threshold $\tau_t$ as the three-sigma bound $\tau_t \, {=} \, 3\,\sigma_t$, which we use in the next section for the data association search.

\subsection{Step~4: Alignment Through Robust Optimization}
\label{sec:optimization}

We base our registration on classic point-to-point~\ac*{icp}~\cite{besl1992pami}. The advantage of this choice is that we do not need to compute data-dependent features such as normals, curvature, or other descriptors, which may depend on the scanner or the environment. Furthermore, with noisy or sparse LiDAR scanners, features such as normals are often not very reliable. Thus, neglecting quantities such as normals in the alignment process is an explicit design decision that allows our system to generalize well to different sensor resolutions.

To obtain the global estimation of the pose~$\mq{T}_t$ of the robot, we start by applying our prediction model $\Tcv$ to the scan~$\set{\hat{P}^*}$ in the local frame. Successively, we transform it into the global coordinate frame using the previous pose estimate~$\mq{T}_{t-1}$, resulting in the source points
\begin{align}
	\label{eq:optimization0}
	\set{S} = \left\{ \d{s}_i = \mq{T}_{t-1} \Tcv \, \d{p} \mid \d{p} \in \set{\hat{P}}^* \right\}.
\end{align}

For each iteration $j$ of~\ac*{icp}, we obtain a set of correspondences between the point cloud $ \set{S} $ and the local map~$\set{Q} \,{=}\, \{ \d{q}_i \,{\mid}\, \d{q}_i \,{\in}\, \RR^3\}$ through nearest neighbor search over the voxel grid (\secref{sec:target-map}) considering only correspondences with a point-to-point distance below $\tau_t$.
To compute the current pose correction $\deltaT{est}{j}$, we perform a robust optimization minimizing the sum of point-to-point residuals
\begin{align}
	\label{eq:optimization2}
	\deltaT{est}{j} = \argmin_{\mq{T}} \sum_{(s,q) \in \set{C}(\tau_t)} \rho(\|\mq{T}  \d{s} - \d{q}\|_2 ),
\end{align}
where~$\set{C}(\tau_t)$ is the set of nearest neighbor correspondences with a distance smaller than~$\tau_t$ and~$\rho $ is the Geman-McClure robust kernel,~\ie, an M-estimator with a strong outlier rejection property, given by
\begin{equation}
	\label{eq:kernel}
	\rho(e) =
	\frac{e^2/2}{\underbrace{\sigma_{t}/3}_{\kappa_t} + e^2},
\end{equation}
where the scale parameter $ \kappa_t $ of the kernel is adapted online using $\sigma_t $. Lastly, we update the points $\d{s}_i$,~\ie,
\begin{equation}
	\label{eq:optimization3}
	\left\{ \d{s}_i \gets \Delta \mq{T}_{\text{est},j} \d{s}_i \mid \d{s}_i \in \set{S}\right\},
\end{equation}
and repeat the process until the convergence criterion is met.

As a result of this process, we obtain the transformation \mbox{$\mq{T}_t \, {=} \, \Delta \mq{T}_{\text{icp}, t} \mq{T}_{t-1} \Tcv $}, where $\Delta \mq{T}_{\text{icp}, t} \, {=} \, \prod_{j}\Delta \mq{T}_{\text{est}, j}$. While we apply the prediction model~$\Tcv$ (\ie, the constant velocity prediction) to the local coordinate frame of the scan, we perform the~\ac*{icp} correction $ \Delta \mq{T}_{\text{icp},t}$ in the global reference frame of the robot. This is done for efficiency reasons as it allows us to transform the source points $\set{S}$ only once per~\ac*{icp} iteration. With this, the local pose deviation $\Delta \mq{T}_t$ at time $t$ used in the~\eqref{eq:delta-points} can be expressed as
\begin{equation}
	\label{eq:delta}
	\Delta \mq{T}_t = \left( \mq{T}_{t-1} \Tcv \right)^{-1} \Delta \mq{T}_{\text{icp},t} \mq{T}_{t-1} \Tcv.
\end{equation}

A standard termination criterion for the ICP algorithm is to control the number of iterations. Additionally, most approaches also have a further criterion based on the minimum change in the solution. Conversely, we found that controlling the number of iterations does not allow the algorithm to always find a good solution. Thus, we only employ the termination criterion based on the applied correction being smaller than~$\gamma$, without imposing a maximum number of iterations.

Finally, the ICP correction is applied to the point cloud~$\set{P}^*_\text{merge}$, and the points are integrated into the local map.

\subsection{Parameters}
Our implementation depends on a small set of seven parameters. All are shown in \tabref{tab:params}. We use the same parameters for all experiments.
Most other approaches use a substantially larger set of parameters: \mulls has $107$ parameters, \suma has $49$ parameters, and \cticp has $30$ parameters in their respective configuration files. In contrast, our approach only has two parameters for the correspondence search, four for the map representation and scan subsampling, and one for the ICP termination.
Note that the maximum range of a scanner is a value that depends on the specific sensor in use and, as such, we do not consider it a system parameter. However, for some scenarios, the value of $r_\text{max}$ might also be adapted to the specific environment in which the system is operating, e.g., not considering far away measurements that are usually less accurate.

\begin{table}
	\centering
	\begin{tabularx}{\columnwidth}{Y||Y}
		\toprule
		\textbf{Parameter}                             & \textbf{Value}         \\ \midrule
		Initial threshold $\tau_0$                     & 2\,m                   \\
		Min. deviation threshold $\delta_{\text{min}}$ & 0.1\,m                 \\
		Max. points per voxel $N_{\text{max}}$         & 20                     \\
		Voxel size map $v$                             & $0.01 \, r_\text{max}$ \\
		Factor voxel size map merge $\alpha$           & $0.5$                  \\
		Factor voxel size registration $\beta$         & $1.5$                  \\
		\ac*{icp} convergence criterion $\gamma$       & $10^{-4}$              \\
		\bottomrule
	\end{tabularx}
	\caption{All seven parameters of our approach.}
	\label{tab:params}
	\vspace{-0.5cm}
\end{table}

\section{Experimental Evaluation}
\label{sec:exp}

This work provides a simple yet effective LiDAR odometry pipeline that comes with a small set of parameters. We present our experiments to show the capabilities of our method. The results of our experiments support our key claims, namely that our approach
(i) is on par with more complex state-of-the-art odometry systems,
(ii) can accurately compute the robot's odometry in a large variety of environments and motion profiles with the same system configuration, and
(iii) provides an effective solution to motion distortion without relying on IMUs or wheel odometers.

\subsection{Experimental Setup}
\label{sec:experimental-setup}
We use numerous datasets and common evaluation methods.
We start with the \kitti to evaluate our system against state-of-the-art approaches to LiDAR odometry. To investigate how we perform in other autonomous driving datasets employing a different sensor, we evaluate our approach on the MulRan dataset~\cite{jeong2018icra}. Additionally, we show that our approach can be used in different scenarios, such as the one present in the \nclt, a segway dataset, and the \ncd recorded using a handheld device. We also analyze our method's different components, such as the motion-compensation scheme and the adaptive threshold.

Please note that due to space limitations we omit to show the results of the trajectories and a detailed runtime evaluation in this manuscript but refer the reader to the official project page where all the plots and per-sequence evaluation on the runtime performances are available.\footnote{\url{https://www.github.com/PRBonn/kiss-icp/tree/main/evaluation}}

\subsection{Performance on the KITTI-Odometry Benchmark}
\label{sec:kitti-benchmark}
This experiment evaluates the performance of different odometry pipelines on the popular KITTI benchmark dataset. Since most systems do not do motion compensation, we use the already compensated KITTI scans for a fair comparison and disable the motion compensation for our approach and \cticp in this first analysis~(the performance of the motion compensation module will be studied later in \secref{sec:ablation-motion-compensation}).
\tabref{tab:kitti} exhibits how our system challenges most state-of-the-art systems, which are typically more sophisticated than our point-to-point~\ac*{icp}. Based on the official KITTI Benchmark, we rank second among the open-source approaches (behind \cticp) and ninth among all submissions. This indicates that our comparably simple system still performs better than all the publicly available systems out there, except \cticp. Note that CT-ICP is a complete~\ac*{slam} system, and it uses loop closures to correct for the accumulated drift of the odometry estimation. We in contrast obtain our results using only open-loop registration without any loop closing.

\begin{table}[t]
	\centering
	\begin{tabularx}{\columnwidth}{Y|c||Y|Y}
		\toprule
		                                                        & \textbf{Method   } & \textbf{Seq. 00-10 } & \textbf{Seq. 11-21} \\ \midrule
		\multirow{3}{*}{\begin{sideways}SLAM\end{sideways}}
		                                                        & \sumapp            & \N 0.70              & \N 1.06             \\
		                                                        & \mulls             & \N 0.52              & \N -                \\
		                                                        & \cticp             & \N 0.53              & \B 0.59             \\
		\bottomrule
		\toprule
		\multirow{5}{*}{\begin{sideways}Odometry\end{sideways}} & \imls              & \N 0.55              & \N 0.69             \\
		                                                        & \mulls             & \N 0.55              & \N 0.65             \\
		                                                        & \loam              & \N 0.84              & \N 1.87             \\
		                                                        & \suma              & \N 0.80              & \N 1.39             \\
		                                                        & \ours              & \B 0.50              & \N 0.61             \\
		\bottomrule
	\end{tabularx}
	\caption{KITTI Benchmark results with motion compensated data. We report the average relative translational error in $\%$~\cite{geiger2013ijrr}. We compare across~\ac*{slam} methods employing pose-graph optimization for improved results \textit{(top)} and odometry methods \textit{(bottom)}. We omit the relative rotational error, but these results are available at \url{https://www.cvlibs.net/datasets/kitti/eval_odometry.php}}
	\label{tab:kitti}
	\vspace{-0.3cm}
\end{table}

\subsection{Comparison to State-of-the-Art Systems on Other Datasets}
We proceed to analyze the performance of our system on different datasets, scenarios, and types of robots. For that, we use the \mulran, a handheld device~\cite{ramezani2020iros}, and a segway dataset~\cite{carlevaris-bianco2016ijrr}. Odometry pipelines typically deal with those challenging scenarios but employ IMUs~\cite{shan2020iros} or a different system configuration~\cite{dellenbach2022icra}. Our system performs on par with state-of-the-art systems using the same parameter values for all experiments and datasets.
For this experiment, we compare against state-of-the-art odometry systems, namely~\mulls, \suma, \loam, and \cticp. Note that we do not provide an evaluation of CT-ICP for the MulRan dataset since CT-ICP does not provide support for this dataset.

For the \mulran, we test the systems under evaluation on all available public sequences. Since the dataset provides three similar runs for each sequence, we report the average number of each sequence in~\tabref{tab:mulran}. Our method outperforms all state-of-the-art approaches by a large margin in both relative and absolute error.

We use both available sequences to evaluate the Newer College dataset and achieve similar results on the short experiment compared to~\ac*{cticp}. For the long experiment, the performance gap can be explained by the additional loop closing module of~\ac*{cticp},  which is a complete~\ac*{slam} system. For the NCLT dataset experiment, we use the sequence evaluated on the original work of CT-ICP. We could not reproduce the results reported in \cticp and therefore report the results given in the original paper~\cite{dellenbach2022icra} in~\tabref{tab:ncd}. We achieve similar results than~\ac*{cticp}. However, we observed errors in the GPS ground-truth poses and missing frames. Therefore, the numbers on NCLT should be taken with a grain of salt and rather provide an estimate of how the systems perform.
We discourage using NCLT to evaluate odometry systems: misalignments in the ground truth poses, missing frames, and inconsistencies in the data make the evaluation of odometry systems on such a dataset not a good evaluation tool from our perspective. However, we provide the results for completeness.

We show qualitative results in~\figref{fig:motivation} generated using our KISS-ICP poses. Using a single system configuration, we can produce consistent maps on different sensor setups (Velodyne/Ouster vs. Livox) and different motion profiles (car, drone, segway, handheld) with the same parameters.

\begin{table}[t]
	\begin{tabularx}{\columnwidth}{c||c|Y|Y|Y|Y}
		\toprule
		\textbf{Sequence}          & \textbf{Method} & \textbf{\makecell{Avg.                                  \\tra.}}&\textbf{\makecell{Avg.\\rot.}}&\textbf{\makecell{ATE\\tra.}}&\textbf{\makecell{ATE\\rot.}}\\
		\midrule
		\multirow{4}{*}{KAIST}     & \mulls          & \N 2.94                & \N 0.86 & \N 37.24   & \N 0.11 \\
		                           & \suma           & \N 5.59                & \N 1.73 & \N 43.61   & \N 0.14 \\
		                           & \loam           & \N 3.43                & \N 0.99 & \N 46.17   & \N 0.15 \\
		                           & \ours           & \B 2.28                & \B 0.68 & \B 17.40   & \B 0.06 \\

		\midrule
		\multirow{4}{*}{DCC}       & \mulls          & \N 2.96                & \N 0.98 & \N 38.35   & \N 0.12 \\
		                           & \suma           & \N 5.20                & \N 1.71 & \N 36.22   & \N 0.11 \\
		                           & \loam           & \N 3.83                & \N 1.14 & \N 42.70   & \N 0.13 \\
		                           & \ours           & \B 2.34                & \B 0.64 & \B 15.16   & \B 0.05 \\

		\midrule
		\multirow{4}{*}{Riverside} & \mulls          & \N 5.42                & \N 2.21 & \N 91.16   & \N 0.16 \\
		                           & \suma           & \N 13.86               & \N 2.13 & \N 227.24  & \N 0.38 \\
		                           & \loam           & \N 5.47                & \N 1.18 & \N 138.09  & \N 0.22 \\
		                           & \ours           & \B 2.89                & \B 0.64 & \B 49.02   & \B 0.08 \\

		\midrule
		\multirow{4}{*}{Sejong*}   & \mulls          & \N 5.93                & \N 0.84 & \N 2151.00 & \N 0.49 \\
		                           & \loam           & \N 7.87                & \N 1.20 & \N 3448.97 & \N 0.82 \\
		                           & \ours           & \B 4.69                & \B 0.70 & \B 1369.54 & \B 0.33 \\
		\bottomrule
	\end{tabularx}
	\caption{Quantitative results on the \mulran. We report the relative translational error and the relative rotational error using the KITTI~\cite{geiger2013ijrr} metrics. Additionally, we show the absolute trajectory error for translation in\,m and for rotation in~$\mathrm{rad}$.}
	\label{tab:mulran}
\end{table}

\begin{table}
	\begin{tabularx}{\columnwidth}{Y||Y|Y||Y}
		\toprule
		\textbf{Method} & \textbf{\makecell{NCD                     \\01-short}}	&\textbf{\makecell{NCD\\02-long}}		& \textbf{\makecell{NCLT\\2012-01-8}} \\
		\midrule
		\mulls          & \N 0.82               & \N 1.23 & -       \\
		\loam           & \N 2.02               & fails   & -       \\
		\cticp          & \B 0.48               & \B 0.58 & \B 1.17 \\
		\ours           & \N 0.51               & \N 0.96 & \N 1.27 \\
		\bottomrule
	\end{tabularx}
	\caption{Quantitative results for Newer College and NCLT. We report the relative translational error in $\%$~\cite{geiger2013ijrr}.}
	\label{tab:ncd}
\end{table}

\subsection{Ablation Studies}
To understand how each component of our system impacts the odometry performance, we conduct ablation studies on the different components of our approach, namely, the motion compensation scheme and the adaptive threshold. To carry out these studies, we use the \kitti as it is probably the best-known one.

\subsubsection{Motion Compensation}
\label{sec:ablation-motion-compensation}
To assess the impact of our motion compensation scheme, we utilize the raw LiDAR point clouds without any compensation applied. Note that the KITTI odometry benchmark point cloud data~\cite{geiger2012cvpr} is already compensated and, therefore, cannot be used for this study. Thus, we use the KITTI raw dataset~\cite{geiger2013ijrr}. We present the results in a familiar fashion, selecting only the sequences that correspond to the ones on the motion-compensated datasets~\cite{geiger2012cvpr}. As we can see in \tabref{tab:deskew}, our motion compensation scheme can produce state-of-the-art results and is on par with substantially more sophisticated and thus complex compensation techniques such as the one introduced by \cticp. Additionally, we study how our system performs without applying motion compensation, as shown in~\tabref{tab:deskew}. We also evaluate the performance of our constant velocity model for motion compensation. To assess this, we compare the same compensation strategy but replace the velocity estimation with sensor data taken by the IMU. As seen in the results, our velocity estimation is on par or even slightly better with the IMU.

Besides the fact that CT-ICP's elastic formulation yields good results, our much simpler approach produces even better results. This result shows that the constant velocity model employed in our approach for compensating motion distortion is sufficient to cope with the slight reduction in performance when no compensation is applied. Consequently, we believe that more sophisticated techniques are unnecessary for \emph{most} robotic odometry estimation.

\begin{table}
	\centering
	\begin{tabularx}{\columnwidth}{c||Y|Y|Y}
		\toprule
		\textbf{Method}           & \textbf{Avg. tra} & \textbf{Avg. rot} & \textbf{Avg. freq.} \\
		\midrule
		\mulls                    & \N 1.41           & -                 & 12\,Hz              \\
		\imls                     & \N 0.71           & -                 & 1\,Hz               \\
		\cticp                    & \N 0.55           & -                 & 15\,Hz              \\
		\ours{} without deskewing & \N 0.91           & 0.27              & 51\,Hz              \\
		\oursimu                  & \N 0.51           & 0.19              & 38\,Hz              \\
		\oursds                   & \B 0.49           & 0.16              & 38\,Hz              \\
		\bottomrule
	\end{tabularx}
	\caption{Results of evaluating different state-of-the-art systems on KITTI-raw dataset~(without motion compensation). We report the relative translational error and the relative rotational error using the KITTI~\cite{geiger2013ijrr} metrics. Additionally, we report the runtime operation of the systems being in consideration for this experiment.}
	\label{tab:deskew}
\end{table}

\begin{table}
	\centering
	\begin{tabularx}{\columnwidth}{c||Y|Y|Y|Y|Y}
		\toprule
		\multirow{2}{*}{\textbf{Dataset}} & \multicolumn{5}{c}{\textbf{Data-Association Threshold~$\tau$}}                                         \\
		                                  & {0.3\,m}                                                       & 0.5\,m  & 1.0\,m  & 2.0\,m  & Ours    \\
		\midrule
		KITTI Seq. 00                     & \N 0.54                                                        & \B 0.51 & \N 0.53 & \N 0.55 & \B 0.51 \\
		KITTI Seq. 04                     & \N 0.39                                                        & \N 0.41 & \N 0.37 & \N 0.39 & \B 0.36 \\
		\midrule
		KITTI Avg. Seq. 00-10             & \N 0.53                                                        & \N 0.51 & \N 0.51 & \N 0.53 & \B 0.50 \\
		\bottomrule
	\end{tabularx}
	\caption{Comparison of different fixed thresholds vs.~our proposed adaptive threshold on the KITTI dataset. We report the relative translational error in $\%$~\cite{geiger2013ijrr}.}
	\label{tab:threshold}
\end{table}

\subsubsection{Adaptive Data-Association Threshold}
We finally evaluate how the adaptive threshold~$\tau_t$ impacts the performance of our system by comparing it to a different set of fixed thresholds commonly used in open-source systems. To conduct this experiment, we identify the two KITTI sequences with the largest (00) and the smallest (04) average acceleration indicating different motion profiles. As we can see in \tabref{tab:threshold}, the best fixed threshold for sequence 00 is~0.5\,m and~1.0\,m for sequence 04. This means that a fixed threshold has to be tuned depending on the motion profile and thus to the dataset to achieve top performance. In contrast, our adaptive algorithm exploits the motion profile to estimate the threshold online, which results in on-par or better performance without the need to find a new fixed threshold for each sequence. Finally, our proposed adaptive threshold strategy achieves the best average result on the KITTI training sequences.

Please note that all the experiments from this ablation study use the robust kernel. For space reasons, we omitted the results of the evaluation of our system when no kernel is employed and only report the results averaged over the sequences. Not using the kernel produces $0.67$\% for the translational error and $0.25$\% for the rotational error~\cite{geiger2013ijrr}.

\section{Conclusion}
\label{sec:conclusion}
This paper presents a simple yet highly effective approach to LiDAR odometry and shows that point-to-point~\ac*{icp} works very well -- when used properly. Our approach operates solely on point clouds and does not require an IMU, even when dealing with high-frequency driving profiles. Our approach exploits the classical point-to-point~\ac*{icp} to build a generic odometry system that can be employed in different challenging environments, such as highway runs, handheld devices, segways, and drones. Moreover, the system can be used with different range-sensing technologies and scanning patterns. We only assume that point clouds are generated sequentially as the robot moves through the environment. We implemented and evaluated our approach on different datasets, provided comparisons to other existing techniques, supported all claims made in this paper, and released our code. The experiments suggest that our approach is on par with substantially more sophisticated state-of-the-art LiDAR odometry systems but relies only on a few parameters, and performs well on various datasets under different conditions with the same parameter set. Finally, our system operates faster than the sensor frame rate in all presented datasets. We believe this work will be a new baseline for future sensor odometry systems and a solid, high-performance starting point for future approaches. Our open-source code is robust and simple, easy to extend, and performs well, pushing the state-of-the-art LiDAR odometry to its limits and challenging most sophisticated systems.

\section{Acknowledgements}
\label{sec:wcknowledgements}
We thank Pierre Dellenbach for making his CT-ICP code available, which inspired our implementation. We thank Yue Pan for helping with the evaluation of MULLS for this paper. Thanks also to Igor Bogoslavskyi for his feedback.

\bibliographystyle{plain_abbrv}
\bibliography{bibexport}
\end{document}

%% file: stachnisslab-latex.tex
\usepackage{graphics}           
\usepackage{times}              
\usepackage{amsmath}            
\usepackage{amssymb}            
\usepackage{graphicx}
\usepackage{algorithm}
\usepackage[noend]{algpseudocode}
\usepackage{booktabs}
\usepackage{color}
\definecolor{instructioncolor}{rgb}{.5,.5,.5}

\usepackage[font=small]{caption}

\def\secref#1{Sec.~\ref{#1}}
\def\figref#1{Fig.~\ref{#1}}
\def\tabref#1{Tab.~\ref{#1}}
\def\eqref#1{Eq.~(\ref{#1})}

\makeatletter
\usepackage{xspace}
\DeclareRobustCommand\onedot{\futurelet\@let@token\@onedot}
\def\@onedot{\ifx\@let@token.\else.\null\fi\xspace}
 
\def\ie{i.e\onedot}

\def\etal{{et al}\onedot}
\makeatother

\def\etalcite#1{\etal~\cite{#1}}

\usepackage{array}
\newcolumntype{L}[1]{>{\raggedright\let\newline\\\arraybackslash\hspace{0pt}}m{#1}}
\newcolumntype{C}[1]{>{\centering\let\newline\\\arraybackslash\hspace{0pt}}m{#1}}
\newcolumntype{R}[1]{>{\raggedleft\let\newline\\\arraybackslash\hspace{0pt}}m{#1}}

%% file: stachnisslab-math.tex
\def\argmin{\mathop{\rm argmin}}

\newcommand{\RR}{\mathbb{R}}

\renewcommand{\b}[1]{\mbox{\boldmath$#1$}}

\renewcommand{\d}[1]{\b {#1}}

\newcommand{\m}[1]{{\mbox{{\sffamily\slshape{#1\/}}}}}

\newcommand{\mq}[1]{{\mbox{{\sffamily{#1}}}}}

\newcommand{\rot}[2]{\m{R}_{#1}(#2)}



%% file: stachnisslab-custom.tex
\usepackage[acronyms, shortcuts]{glossaries}
\usepackage[normalem]{ulem}
\usepackage{figsize}
\usepackage{float}
\usepackage{multicol}
\usepackage{multirow}
\usepackage{ragged2e}
\usepackage{stmaryrd}
\usepackage{subfigure}
\usepackage{tabularx}
\usepackage{xcolor}
\usepackage{makecell}
\usepackage{rotating}

\newcommand{\ours}{Ours}
\newcommand{\oursds}{Ours\,{+}\,Deskewing (CV)}
\newcommand{\oursimu}{Ours\,{+}\,Deskewing (IMU)}
\newcommand{\B}{\fontseries{b}\selectfont}
\newcommand{\N}{}

\newcommand{\set}[1]{\mathcal{#1}}
\newcommand{\deltaT}[2]{\Delta \mq{T}_{\text{#1},#2}}

\newcommand{\Tcv}{\mq{T}_{\text{pred},t}}

\DeclareMathOperator{\Tr}{tr}

\graphicspath{{pics/}}
\newcolumntype{Y}{>{\centering\arraybackslash}X}


%% file: stachnisslab-glossary.tex
\newacronym{mos}{MOS}{moving object segmentation}
\newacronym{icp}{ICP}{iterative closest point}
\newacronym{slam}{SLAM}{simultaneous localization and mapping}
\newacronym{gicp}{GICP}{Generalized ICP}
\newacronym{loam}{LOAM}{lidar odometry and mapping}
\newacronym{suma}{SuMa}{Surfel-based Mapping}
\newacronym{cticp}{CT-ICP}{continuous time ICP}
\newacronym{imu}{IMU}{inertial measurement unit}

\newcommand{\rgbd}{\mbox{RGB-D}\xspace}

\newcommand{\imls}{IMLS-SLAM~\cite{deschaud2018icra}\xspace}
\newcommand{\mulls}{MULLS~\cite{pan2021icra-mvls}\xspace}
\newcommand{\cticp}{CT-ICP~\cite{dellenbach2022icra}\xspace}
\newcommand{\suma}{SuMa~\cite{behley2018rss}\xspace}
\newcommand{\sumapp}{SuMa\texttt{++}\xspace~\cite{behley2018rss}}
\newcommand{\loam}{F-LOAM~\cite{wang2021iros-fflo}\xspace}

\newcommand{\kitti}{KITTI odometry dataset~\cite{geiger2012cvpr}\xspace}
\newcommand{\mulran}{MulRan dataset~\cite{jeong2018icra}\xspace}
\newcommand{\nclt}{NCLT dataset~\cite{carlevaris-bianco2016ijrr}\xspace}
\newcommand{\ncd}{Newer College dataset~\cite{ramezani2020iros}\xspace}

%% file: pics/adaptive_rot.pdf_tex
\begingroup%
  \makeatletter%
  \providecommand\color[2][]{%
    \errmessage{(Inkscape) Color is used for the text in Inkscape, but the package 'color.sty' is not loaded}%
    \renewcommand\color[2][]{}%
  }%
  \providecommand\transparent[1]{%
    \errmessage{(Inkscape) Transparency is used (non-zero) for the text in Inkscape, but the package 'transparent.sty' is not loaded}%
    \renewcommand\transparent[1]{}%
  }%
  \providecommand\rotatebox[2]{#2}%
  \newcommand*\fsize{\dimexpr\f@size pt\relax}%
  \newcommand*\lineheight[1]{\fontsize{\fsize}{#1\fsize}\selectfont}%
  \ifx\svgwidth\undefined%
    \setlength{\unitlength}{398.7003789bp}%
    \ifx\svgscale\undefined%
      \relax%
    \else%
      \setlength{\unitlength}{\unitlength * \real{\svgscale}}%
    \fi%
  \else%
    \setlength{\unitlength}{\svgwidth}%
  \fi%
  \global\let\svgwidth\undefined%
  \global\let\svgscale\undefined%
  \makeatother%
  \begin{picture}(1,0.63329054)%
    \lineheight{1}%
    \setlength\tabcolsep{0pt}%
    \put(0,0){\includegraphics[width=\unitlength,page=1]{adaptive_rot.pdf}}%
    \put(0.11453747,0.09075417){\makebox(0,0)[lt]{\lineheight{1.25}\smash{\begin{tabular}[t]{l}$\theta$\end{tabular}}}}%
    \put(0.54428868,0.09178414){\makebox(0,0)[lt]{\lineheight{1.25}\smash{\begin{tabular}[t]{l}$r_\text{max}$\end{tabular}}}}%
    \put(0.76741346,0.07032828){\makebox(0,0)[lt]{\lineheight{1.25}\smash{\begin{tabular}[t]{l}$\d{p}$\end{tabular}}}}%
    \put(0.8046019,0.21593709){\makebox(0,0)[lt]{\lineheight{1.25}\smash{\begin{tabular}[t]{l}$\Delta \m{R}$\end{tabular}}}}%
    \put(0.74541388,0.43184627){\makebox(0,0)[lt]{\lineheight{1.25}\smash{\begin{tabular}[t]{l}$\Delta \d{t}$\end{tabular}}}}%
    \put(0.34087947,0.26682549){\rotatebox{28.529829}{\makebox(0,0)[lt]{\lineheight{1.25}\smash{\begin{tabular}[t]{l}$r_\text{max}$\end{tabular}}}}}%
    \put(0.82994805,0.52294152){\makebox(0,0)[lt]{\lineheight{1.25}\smash{\begin{tabular}[t]{l}$\Delta \m{R}\d{p}+\Delta \d{t}$\end{tabular}}}}%
    \put(0.25937874,0.00591521){\color[rgb]{0,0,0}\makebox(0,0)[lt]{\lineheight{1.25}\smash{\begin{tabular}[t]{l}$\delta_\text{trans}(\Delta \d{t})$\end{tabular}}}}%
    \put(0.51928655,0.00591521){\color[rgb]{0,0,0}\makebox(0,0)[lt]{\lineheight{1.25}\smash{\begin{tabular}[t]{l}$\delta_\text{rot}(\Delta \m{R})$\end{tabular}}}}%
    \put(0.30456915,0.39427182){\color[rgb]{0,0,0}\rotatebox{48.853704}{\makebox(0,0)[lt]{\lineheight{1.25}\smash{\begin{tabular}[t]{l}$\delta(\Delta \mq{T})$\end{tabular}}}}}%
  \end{picture}%
\endgroup%

%% file: vizzo2023ral.bbl
\begin{thebibliography}{10}

\bibitem{behley2018rss}
J.~Behley and C.~Stachniss.
\newblock {Efficient Surfel-Based SLAM using 3D Laser Range Data in Urban
  Environments}.
\newblock In {\em Proc.~of Robotics: Science and Systems (RSS)}, 2018.

\bibitem{bentley1975kdtree}
J.~Bentley.
\newblock Multidimensional binary search trees used for associative searching.
\newblock {\em Communications of the ACM}, 18(9):509--517, 1975.

\bibitem{besl1992pami}
P.~Besl and N.~McKay.
\newblock {A Method for Registration of 3D Shapes}.
\newblock {\em IEEE Trans.~on Pattern Analalysis and Machine Intelligence
  (TPAMI)}, 14(2):239--256, 1992.

\bibitem{blanco2014mrpt}
J.L. Blanco-Claraco.
\newblock Mobile robot programming toolkit (mrpt).
\newblock {\em URL: http://www. mrpt. org/}, 2014.

\bibitem{carlevaris-bianco2016ijrr}
N.~Carlevaris-Bianco, A.~Ushani, and R.~Eustice.
\newblock {University of Michigan North Campus long-term vision and lidar
  dataset}.
\newblock {\em Intl.~Journal~of Robotics Research (IJRR)}, 35(9):1023--1035,
  2016.

\bibitem{chebrolu2021ral}
N.~Chebrolu, T.~L\"{a}be, O.~Vysotska, J.~Behley, and C.~Stachniss.
\newblock {Adaptive Robust Kernels for Non-Linear Least Squares Problems}.
\newblock {\em IEEE Robotics and Automation Letters (RA-L)}, 6:2240--2247,
  2021.

\bibitem{chen2021ral}
X.~Chen, S.~Li, B.~Mersch, L.~Wiesmann, J.~Gall, J.~Behley, and C.~Stachniss.
\newblock {Moving Object Segmentation in 3D LiDAR Data: A Learning-based
  Approach Exploiting Sequential Data}.
\newblock {\em IEEE Robotics and Automation Letters (RA-L)}, 6:6529--6536,
  2021.

\bibitem{chen2019iros}
X.~Chen, A.~Milioto, E.~Palazzolo, P.~Giguère, J.~Behley, and C.~Stachniss.
\newblock {SuMa++: Efficient LiDAR-based Semantic SLAM}.
\newblock In {\em Proc.~of the IEEE/RSJ Intl.~Conf.~on Intelligent Robots and
  Systems (IROS)}, 2019.

\bibitem{chen1991iros}
Y.~Chen and G.~Medioni.
\newblock Object modeling by registration of multiple range images.
\newblock In {\em Proc.~of the IEEE/RSJ Intl.~Conf.~on Intelligent Robots and
  Systems (IROS)}, 1991.

\bibitem{dellenbach2022icra}
P.~Dellenbach, J.~Deschaud, B.~Jacquet, and F.~Goulette.
\newblock {CT-ICP Real-Time Elastic LiDAR Odometry with Loop Closure}.
\newblock In {\em Proc.~of the IEEE Intl.~Conf.~on Robotics \& Automation
  (ICRA)}, 2022.

\bibitem{deschaud2018icra}
J.~Deschaud.
\newblock Imls-slam: scan-to-model matching based on 3d data.
\newblock In {\em Proc.~of the IEEE Intl.~Conf.~on Robotics \& Automation
  (ICRA)}, 2018.

\bibitem{geiger2012cvpr}
A.~Geiger, P.~Lenz, and R.~Urtasun.
\newblock {Are we ready for Autonomous Driving? The KITTI Vision Benchmark
  Suite}.
\newblock In {\em Proc.~of the IEEE Conf.~on Computer Vision and Pattern
  Recognition (CVPR)}, 2012.

\bibitem{geiger2013ijrr}
A.~Geiger, P.~Lenz, C.~Stiller, and R.~Urtasun.
\newblock {Vision meets Robotics: The KITTI Dataset}.
\newblock {\em Intl.~Journal~of Robotics Research (IJRR)}, 32(11), 2013.

\bibitem{guadagnino2022ral}
T.~Guadagnino, X.~Chen, M.~Sodano, J.~Behley, G.~Grisetti, and C.~Stachniss.
\newblock {Fast Sparse LiDAR Odometry Using Self-Supervised Feature Selection
  on Intensity Images}.
\newblock {\em IEEE Robotics and Automation Letters (RA-L)}, 7(3):7597--7604,
  2022.

\bibitem{jeong2018icra}
J.~Jeong, Y.~Cho, Y.~Shin, H.~Roh, and A.~Kim.
\newblock Complex urban lidar data set.
\newblock In {\em Proc.~of the IEEE Intl.~Conf.~on Robotics \& Automation
  (ICRA)}, 2018.

\bibitem{kolluri2008talg}
R.~Kolluri.
\newblock Provably good moving least squares.
\newblock {\em ACM Transactions on Algorithms (TALG)}, 4(2):1--25, 2008.

\bibitem{lin2019iros-larl}
J.~Lin and F.~Zhang.
\newblock {Loam\_livox A Robust LiDAR Odemetry and Mapping LOAM Package for
  Livox LiDAR}.
\newblock In {\em Proc.~of the IEEE/RSJ Intl.~Conf.~on Intelligent Robots and
  Systems (IROS)}, 2019.

\bibitem{museth2013siggraph}
K.~Museth, J.~Lait, J.~Johanson, J.~Budsberg, R.~Henderson, M.~Alden, P.~Cucka,
  D.~Hill, and A.~Pearce.
\newblock Openvdb: an open-source data structure and toolkit for
  high-resolution volumes.
\newblock In {\em ACM SIGGRAPH Courses}. 2013.

\bibitem{newcombe2011ismar}
R.A. Newcombe, S.~Izadi, O.~Hilliges, D.~Molyneaux, D.~Kim, A.J. Davison,
  P.~Kohli, J.~Shotton, S.~Hodges, and A.~Fitzgibbon.
\newblock {KinectFusion: Real-Time Dense Surface Mapping and Tracking}.
\newblock In {\em Proc.~of the Intl.~Symposium~on Mixed and Augmented Reality
  (ISMAR)}, 2011.

\bibitem{niessner2013siggraph}
M.~Nie{\ss}ner, M.~Zollh{\"o}fer, S.~Izadi, and M.~Stamminger.
\newblock {Real-time 3D Reconstruction at Scale using Voxel Hashing}.
\newblock In {\em Proc.~of the SIGGRAPH Asia}, 2013.

\bibitem{pan2021icra-mvls}
Y.~Pan, P.~Xiao, Y.~He, Z.~Shao, and Z.~Li.
\newblock {MULLS: Versatile LiDAR SLAM Via Multi-Metric Linear Least Square}.
\newblock In {\em Proc.~of the IEEE Intl.~Conf.~on Robotics \& Automation
  (ICRA)}, 2021.

\bibitem{ramezani2020iros}
M.~{Ramezani}, Y.~{Wang}, M.~{Camurri}, D.~{Wisth}, M.~{Mattamala}, and
  M.~{Fallon}.
\newblock The newer college dataset: Handheld lidar, inertial and vision with
  ground truth.
\newblock In {\em Proc.~of the IEEE/RSJ Intl.~Conf.~on Intelligent Robots and
  Systems (IROS)}, 2020.

\bibitem{reijgwart2019ral}
V.~Reijgwart, A.~Millane, H.~Oleynikova, R.~Siegwart, C.~Cadena, and J.~Nieto.
\newblock Voxgraph: Globally consistent, volumetric mapping using signed
  distance function submaps.
\newblock {\em IEEE Robotics and Automation Letters (RA-L)}, 5(1):227--234,
  2019.

\bibitem{rusinkiewicz2001dim}
S.~Rusinkiewicz and M.~Levoy.
\newblock {Efficient variants of the ICP algorithm}.
\newblock In {\em Proc.~of Int. Conf. on 3-D Digital Imaging and Modeling},
  2001.

\bibitem{rusu2011icra}
R.B. Rusu and S.~Cousins.
\newblock 3d is here: Point cloud library (pcl).
\newblock In {\em Proc.~of the IEEE Intl.~Conf.~on Robotics \& Automation
  (ICRA)}, 2011.

\bibitem{serafin2015iros}
J.~Serafin and G.~Grisetti.
\newblock {NICP: Dense Normal Based Point Cloud Registration}.
\newblock In {\em Proc.~of the IEEE/RSJ Intl.~Conf.~on Intelligent Robots and
  Systems (IROS)}, pages 742--749, 2015.

\bibitem{shan2020iros}
T.~Shan, B.~Englot, D.~Meyers, W.~Wang, C.~Ratti, and D.~Rus.
\newblock {LIO-SAM Tightly-Coupled Lidar Inertial Odometry Via Smoothing and
  Mapping}.
\newblock In {\em Proc.~of the IEEE/RSJ Intl.~Conf.~on Intelligent Robots and
  Systems (IROS)}, 2020.

\bibitem{shan2018iros}
T.~Shan and B.~Englot.
\newblock {LeGO-LOAM: Lightweight and Ground-Optimized Lidar Odometry and
  Mapping on Variable Terrain}.
\newblock In {\em Proc.~of the IEEE/RSJ Intl.~Conf.~on Intelligent Robots and
  Systems (IROS)}, 2018.

\bibitem{thrun2005probrobbook}
S.~Thrun, W.~Burgard, and D.~Fox.
\newblock {\em {Probabilistic Robotics}}.
\newblock MIT Press, 2005.

\bibitem{vespa2018ral}
E.~Vespa, N.~Nikolov, M.~Grimm, L.~Nardi, P.~Kelly, and S.~Leutenegger.
\newblock Efficient octree-based volumetric slam supporting signed-distance and
  occupancy mapping.
\newblock {\em IEEE Robotics and Automation Letters (RA-L)}, 3(2):1144--1151,
  2018.

\bibitem{vizzo2021icra}
I.~Vizzo, X.~Chen, N.~Chebrolu, J.~Behley, and C.~Stachniss.
\newblock {Poisson Surface Reconstruction for LiDAR Odometry and Mapping}.
\newblock In {\em Proc.~of the IEEE Intl.~Conf.~on Robotics \& Automation
  (ICRA)}, 2021.

\bibitem{vizzo2022sensors}
I.~Vizzo, T.~Guadagnino, J.~Behley, and C.~Stachniss.
\newblock {VDBFusion: Flexible and Efficient TSDF Integration of Range Sensor
  Data}.
\newblock {\em Sensors}, 22(3):1296, 2022.

\bibitem{wang2021iros-fflo}
H.~Wang, C.~Wang, C.~Chen, and L.~Xie.
\newblock {F-LOAM: Fast LiDAR Odometry and Mapping}.
\newblock In {\em Proc.~of the IEEE/RSJ Intl.~Conf.~on Intelligent Robots and
  Systems (IROS)}, 2021.

\bibitem{zeng2013gmodels}
M.~Zeng, F.~Zhao, J.~Zheng, and X.~Liu.
\newblock Octree-based fusion for realtime 3d reconstruction.
\newblock {\em Graphical Models}, 75(3):126--136, 2013.

\bibitem{zhang2014rss}
J.~Zhang and S.~Singh.
\newblock {LOAM: Lidar Odometry and Mapping in Real-time}.
\newblock In {\em Proc.~of Robotics: Science and Systems (RSS)}, 2014.

\bibitem{zhou2018arxiv}
Q.~Zhou, J.~Park, and V.~Koltun.
\newblock {Open3D}: {A} modern library for {3D} data processing.
\newblock {\em arXiv:1801.09847}, 2018.

\end{thebibliography}
